\documentclass[11pt]{article}

\usepackage[preprint]{acl}

\usepackage{times}
\usepackage{latexsym}

\usepackage[T1]{fontenc}
\usepackage[utf8]{inputenc}

\usepackage{microtype}

\usepackage{graphicx}
\usepackage{booktabs}
\usepackage{amsmath}
\usepackage{amssymb}
\usepackage{marvosym}
\usepackage{balance}

\AtBeginDocument{%
\setlength{\abovedisplayskip}{5pt plus 2pt minus 2pt}%
\setlength{\belowdisplayskip}{5pt plus 2pt minus 2pt}%
\setlength{\abovedisplayshortskip}{3pt plus 1pt}%
\setlength{\belowdisplayshortskip}{3pt plus 1pt}}

\newcommand{\hime}{\textsc{HiMe}}
\newcommand{\Sref}[1]{\hyperref[#1]{\S\ref{#1}}}

\title{HiMe: Real-Time Self-Hosted Personal Agent Platform for Health Insights with Wearable Devices}

\author{
 \textbf{Wei Liu\textsuperscript{\footnotesize $^{*\bigstar}$}},
 \textbf{Siya Qi\textsuperscript{\footnotesize $^{*\bigstar}$}},
 \textbf{Linhai Zhang\textsuperscript{\footnotesize $^\bigstar$}},
 \textbf{Lorainne Tudor Car\textsuperscript{\footnotesize $^\bigstar$}},
 \textbf{Yulan He\textsuperscript{\footnotesize $^{\bigstar\spadesuit}$\textsuperscript{\Letter}}},
\\
\\
 \textsuperscript{\footnotesize $^\bigstar$}King's College London,
 \textsuperscript{\footnotesize $^\spadesuit$}The Alan Turing Institute,
\\
 \small{
  \texttt{\{wei.4.liu, yulan.he\}@kcl.ac.uk}
 }
}

\begin{document}
\maketitle
\def\thefootnote{*}\footnotetext{Equal contribution.}\def\thefootnote{\arabic{footnote}}

\begin{abstract}
Traditional approaches to wearable health signal analysis, such as smartwatches, are constrained by rigid analytical frameworks and limited personalisation. The emergence of LLM agents creates a new opportunity for Personal Health Agentic Analysis, where health insights can be generated adaptively and in context. However, currently there is no open-source locally deployable platform capable of processing personal health data in real time while preserving privacy. We present \hime{}, a locally deployable, privacy-first agent platform that is fully compatible with real-time health data ecosystems across a wide range of wearable devices. \hime{} is guided by three design principles. The database is treated as a first-class component. Effectiveness and efficiency are jointly optimised to achieve a low-cost Pareto-optimal balance. Data are processed in real time while the user is modelled over the long term. Together, these principles make it practical for individuals to harness Personal Health Agents for continuous, personalised health monitoring for better wellbeing\footnote{Available at \url{https://github.com/thinkwee/HiMe}}.
\end{abstract}

\section{Introduction}
\label{sec:intro}

\begin{figure}[tbp]
    \centering
    \includegraphics[width=0.99\linewidth]{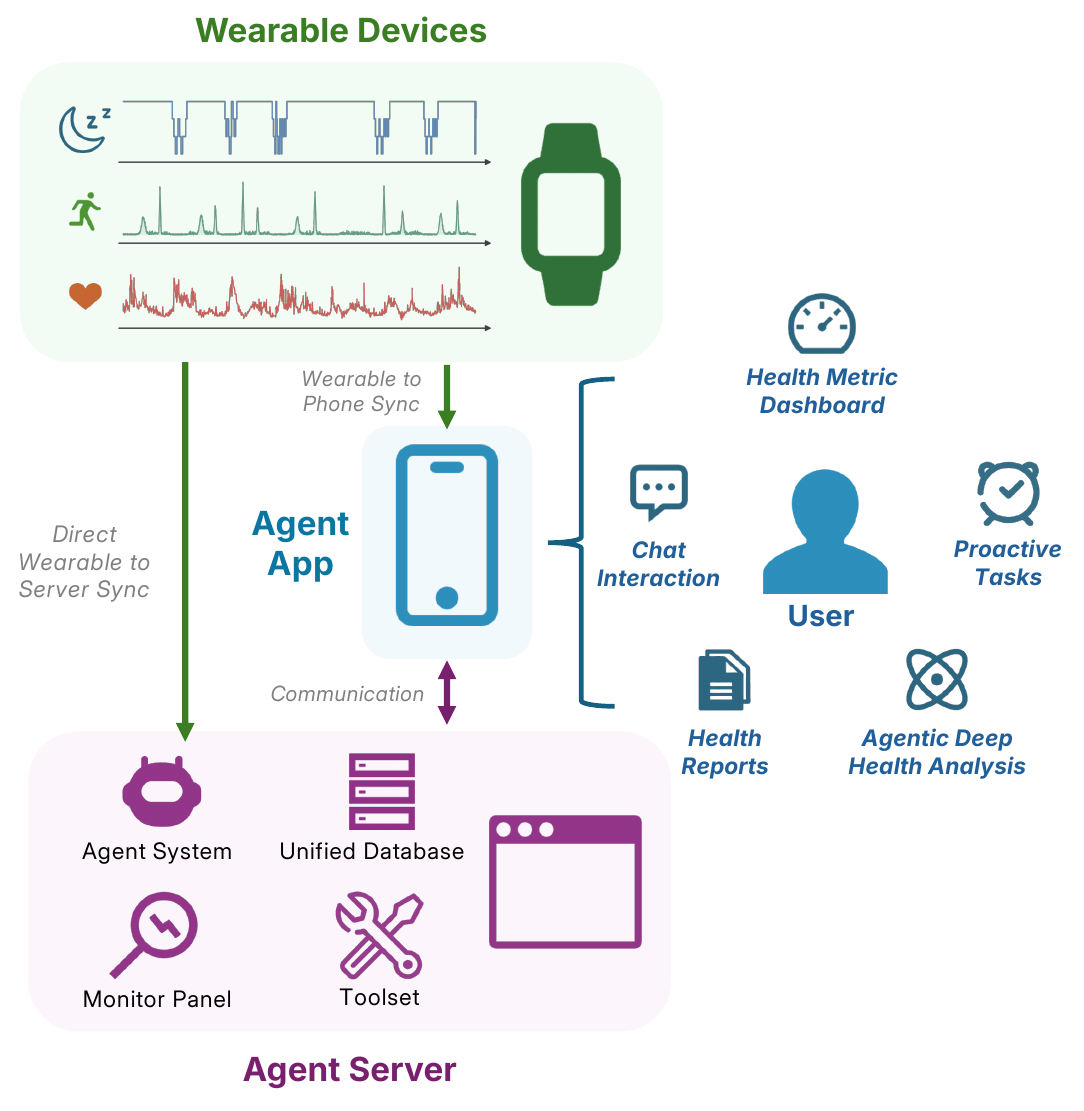}
    \caption{Overview of the \hime{} system.}
    \label{fig:pipeline}
\end{figure}

\begin{figure*}
    \centering
    \includegraphics[width=0.99\linewidth]{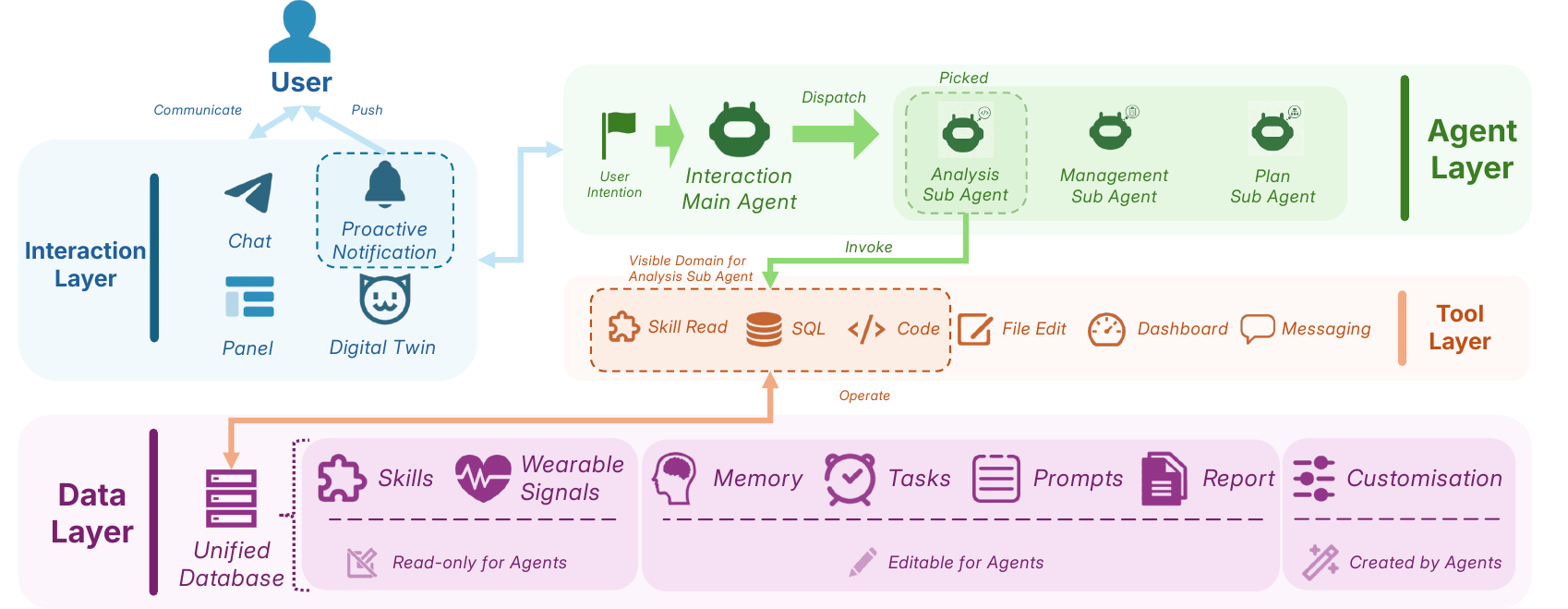}
    \caption{The four layers of the \hime{} framework.}
    \label{fig:architecture}
\end{figure*}

Wearable devices have moved health monitoring out of the clinic and into everyday life. Modern smartwatches and smartphones continuously record signals such as heart rate, blood oxygen, sleep, and workouts into private on-device stores. Over time, these signals form a longitudinal record of an individual's health. But making full use of this stream is hard. Conventional wearable analytics rely on fixed statistics that often ignore both signal dynamics and user preferences, while the long-term raw trace is too large to fit within the context window of a large language model (LLM) directly for intelligent analysis.

LLM agents that reason and act through tools offer a promising alternative~\cite{yao-etal-2023-react}. Such agents can access health data through tools, decide what is worth analysing, and explain their findings in natural language. This promise has mainly been pursued in two settings~\cite{lin2026framework}, neither of which fully addresses individuals living with their own continuous wearable health data. Clinical multi-agent systems reason over encounter-based records for clinician-facing use cases~\citep{tang-etal-2024-medagents,kim-etal-2024-mdagents,schmidgall-etal-2024-agentclinic}, making the clinician, not the patient, the primary user. Personal coaching systems such as GPTCoach~\citep{jorke-etal-2025-gptcoach} and MindScape~\citep{nepal-etal-2024-mindscape} are built around guided dialogue or journal prompts rather than a continuous data stream.
General-purpose self-hosted assistants such as OpenClaw~\citep{steinberger-2026-openclaw} have also popularised a local-first personal agent reached through messaging channels, but are built for chat and scheduled tasks, not a high-rate physiological stream or months-long models of individuals. There is still no open-source platform that lets individuals run, on their own hardware, an agent that ingests wearable data in real time and generates long-term personal health insights.

We present \hime{}, the \textbf{H}ealth \textbf{I}ntelligence \textbf{M}anagement \textbf{E}ngine. This locally deployable, privacy-first agent platform connects to real-time health data ecosystems from a wide range of wearable devices. \hime{} is built on three design principles for personal health.
\textbf{1)} The database is treated as a first-class component. This enables the agent to analyse wearable signals, personal preference and its own memory jointly, while the checkable terminal state of the database provides a direct target for evaluating LLM agents on \hime{}'s health insight generation tasks.
\textbf{2)} Effectiveness and efficiency are optimised jointly, seeking a Pareto-optimal trade-off among quality, cost, and latency, allowing always-on agency to remain affordable on consumer hardware and local models.
\textbf{3)} Real-time data processing is combined with long-term user modelling. This allows the system to capture both momentary anomalies and months-long trends, supporting timely intervention and longitudinal understanding.
Each user is effectively a continuous, heterogeneous data stream. Such personal data are an asset fundamentally different from the public data used to train large models. \hime{} is built to harness LLM agents for this stream, not only to satisfy a momentary request but to help each user become a healthier self over time.
The modular design also supports AI researchers requiring environments for evaluating and training health agents, and medical research applying large models to real-world wearable data.

\section{Architecture}
\label{sec:arch}

\subsection{Overall}
\label{sec:arch:overall}
\hime{} is a self-hosted platform consisting of an agent server and companion apps for the smartphone and wearable (Figure~\ref{fig:pipeline}). A single agent sits between the user and a continuously growing store of personal data, as the user never queries the data directly, and the agent reads, analyses and writes it on the user's behalf, turning a raw physiological stream into personalised insight, with the whole pipeline running on the user's own hardware.
A wearable collects signals such as heart rate, blood oxygen, heart-rate variability, sleep and workouts in real time, buffering readings into batches relayed via the phone to the agent server, where a data adapter normalises any wearable format and deduplicates each sample before storage. All data land in one unified per-user database, a signal store alongside agent-state tables for memory, the user profile, reports, scheduled tasks and monitoring rules, to which the agent may add tables of its own. In the evaluation (\Sref{sec:eval}), public datasets are replayed through the same adapter, so \hime{} doubles as an off-the-shelf platform for wearable agent benchmarking.

The agent server runs a single loop that interleaves user messages, real-time monitor triggers, and scheduled tasks. Each wakes the interaction agent, which routes to specialised sub-agents and acts through a small tool set, reading the signal store over several steps and writing the result back as a report, a memory update or a new task, before replying on the channel the request arrived from. Every reported number is checked against its query evidence. Acting unprompted, a cheap statistical trigger watches the stream at full resolution for free and wakes the expensive analysis only when it fires, pushing the resulting insight to the phone and watch --- turning a continuous physiological stream into personalised, auditable insight without the data ever leaving the user's hardware.

\subsection{Layered Design}
\label{sec:arch:layers}
Figure~\ref{fig:architecture} shows the four layers behind this pipeline.

\paragraph{Data layer.} All information the agent holds lives in one unified database, with tables for wearable signals, memory and the user profile, and the agent may create custom tables for whatever else it needs, so heterogeneous data are analysed jointly. Asked for a health-check report, for example, the agent reads established baselines from its own tables and analysis skills from the skill table, queries the signal table over several steps, replies, writes the result to the report table, and may add a morning proactive task. Each agent holds its own table permissions, so every operation stays safe. No workflow is hard-coded. \hime{} supplies only the tool set and schema design principles rather than rigid constraints, which lets weaker LLMs emit well-formed outputs while leaving stronger ones unconstrained.

\paragraph{Tool layer.} With the database as a first-class component, the tool layer stays simple. In principle, the agent only reads and writes operations over the unified database, and a code execution tool is needed, with scenario-specific variants for replying, editing files or generating dashboards sharing the same abstraction. The code tool keeps Jupyter-style state across calls~\citep{kluyver-etal-2016-jupyter}, so the agent builds on earlier steps instead of rewriting from scratch. To curb tool-use hallucination, a fact verifier classifies every outbound message from its tool trace, checks each reported number against the matching query result, and stores a hashed evidence trail behind a show-evidence button.

\paragraph{Agent layer.} The interaction agent is the primary, user-facing agent, and at any tool-use step it can invoke specialised sub-agents, including an analysis agent, a management agent for memory updates and proactive tasks, and a plan agent for personalised health plans. Prompts are assembled from modular layers (system instructions, role, soul, responsibilities and experience), maximising reuse and keeping the prefix stable for better key-value (KV) cache utilisation. The backend supports 15 LLM API providers, though we recommend serving open-source models locally through vLLM~\citep{kwon-etal-2023-vllm} for privacy.

\paragraph{Interaction layer.} Users reach the agent through the built-in chat or integrated messaging platforms such as WeChat and Telegram, and a gateway registry routes each reply back to the channel the message arrived on. As interactions accumulate, the agent keeps updating its memory and the user profile and configures proactive tasks autonomously, all of which stay editable in the agent server panel, either manually or by instructing the agent. Proactive tasks cover time-based schedules and event-based triggers that the agent authors itself, letting it push notifications to the smartphone and watch. A pixel-art cat acts as a digital twin whose animations reflect the user's current health status in real time. \Sref{sec:demo} walks through a session across these surfaces.

\paragraph{From principles to design.} These three principles, not the field's defaults, explain why the system takes this shape. Treating the database as first-class makes the agent an interface over persistent state rather than a transient reasoning engine. The user model stays as agent-editable Markdown rather than an opaque vector store~\citep{packer-etal-2023-memgpt,singh-etal-2024-mem0,xu-etal-2025-amem}, every action leaves a checkable terminal state that automates evaluation, and a new capability becomes a new table rather than a new agent. Jointly optimising effectiveness and efficiency removes needless computation, as lightweight filters run before expensive inference, narrow roles carry small prompts routed to the cheapest adequate model, and a static prefix improves KV-cache reuse. Pairing real-time processing with long-term modelling lets cheap components watch the stream at full resolution while the LLM wakes only on meaningful triggers, capturing transient anomalies and long-horizon trends in one framework. \Sref{sec:eval} analyses these choices.

\section{Demonstration Walkthrough}
\label{sec:demo}

\begin{figure*}[t]
    \centering
    \includegraphics[width=0.99\linewidth]{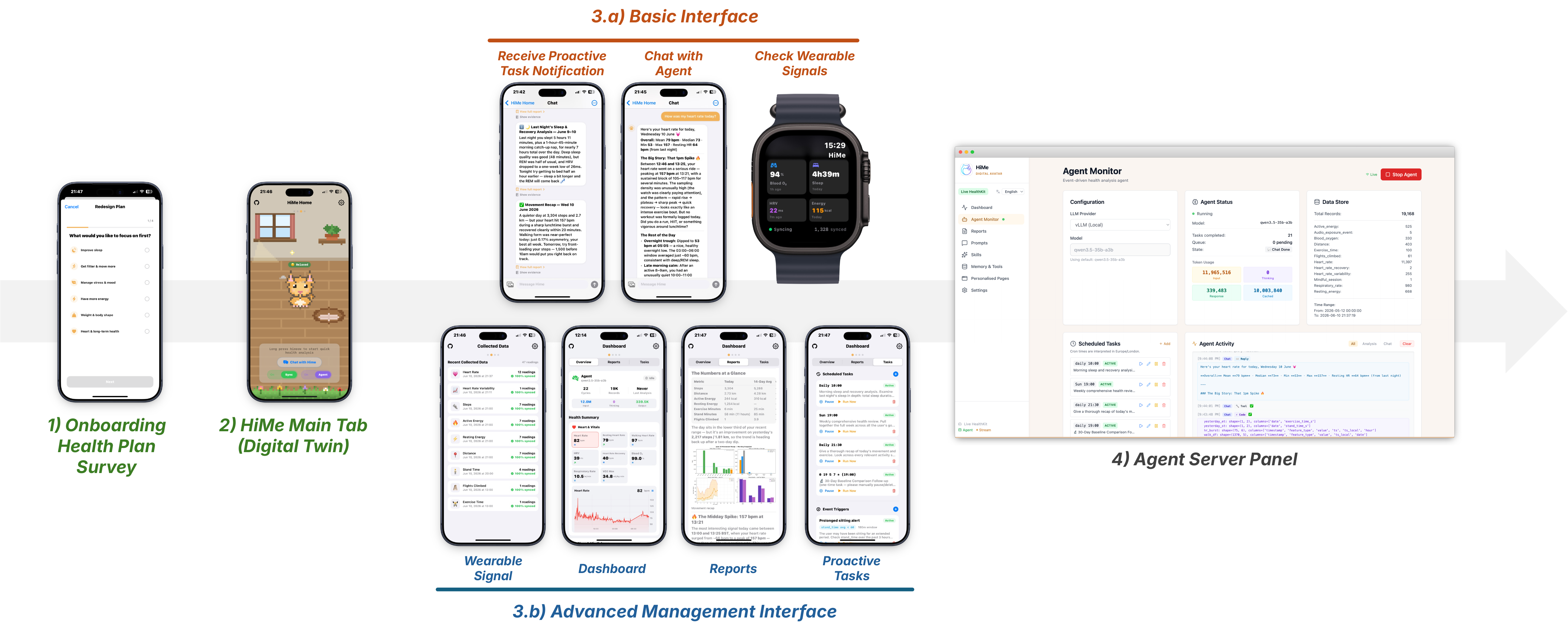}
    \caption{User-facing demonstration flow, from onboarding through a day of use across the in-app chat, wearable device status check, advanced management tabs, and agent server panel for comprehensive monitoring.}
    \label{fig:ui}
\end{figure*}

This section demonstrates a typical user journey from first launch to daily use.\footnote{A screencast of this walkthrough is available at \url{https://www.youtube.com/watch?v=5gurIKyDMH4}.} As shown in Figure~\ref{fig:ui}, deployment begins with a guided setup wizard that configures the LLM backend, timezone, optional messaging integration, and agent server, followed by smartphone onboarding for health-data permission, server connection, consent, and a lightweight goal survey. The plan agent uses these inputs together with available health data to create personalised monitoring tasks, scheduled analyses, and an initial health plan. During daily use, the digital-twin interface provides an ambient summary through the cat avatar, while users can request instant analyses or interact with the agent through chat, notifications, messaging platforms, and wearable views. Agent-generated reports expose supporting evidence through database-backed queries, ensuring transparent and inspectable results (\Sref{sec:arch:layers}). Advanced tabs allow users to inspect collected data, agent status, reports, schedules, and personalised pages, while the agent server panel provides comprehensive control over models, live execution, prompts, memories, skills, tools, and generated content. Together, these interfaces allow users to both benefit from autonomous health assistance and retain full visibility and control over the agent's operation.

\section{Evaluation}
\label{sec:eval}

\hime{} is a demonstration system rather than a benchmark, yet the first principle makes it measurable. Because every capability leaves a terminal database state, most actions are scored by replaying a public wearable trace, letting the agent act and comparing the resulting state against a gold state computed independently from the same trace. We drive the deployed agent offline over ten randomly sampled participants from each of five public wearable corpora, namely LifeSnaps~\citep{yfantidou-etal-2022-lifesnaps}, PMData~\citep{thambawita-etal-2020-pmdata}, MMASH~\citep{rossi-2020-mmash}, CovIdentify~\citep{cho-etal-2024-covidentify} and GLOBEM~\citep{xu-etal-2022-globem}, each through its own data adapter. The sweep covers 22 backbones. Nineteen run locally and span the Qwen2.5~\citep{yang-etal-2024-qwen25}, Qwen3~\citep{yang-etal-2025-qwen3}, Qwen3.5~\citep{qwen-2026-qwen35}, Qwen3.6~\citep{qwen-2026-qwen36-27b} and Gemma-4~\citep{deepmind-2026-gemma4} families from 1.5B to 35B parameters, along with nemotron-3-nano, granite-4.1-8b and glm-4.7-flash. GPT-5-mini, DeepSeek-V4-flash and Gemini-3.5-flash are hosted-API frontier references. On-device execution is the right target for privacy and latency, so the question that matters is how far local small backbones remain from frontier models. We benchmarked latency and show the results in Appendix~\ref{app:latency}.

\subsection{Per-role agentic capability}
\label{sec:eval:roles}

We give every agent in \hime{} the same backbone and score it against a terminal-state oracle that requires no LLM judge. Results are reported as pass@1~\citep{chen-etal-2021-codex} over three runs. Each of the five role columns isolates one capability with its own closed-form check. \textbf{Route} scores top-level dispatch. The interaction agent must hand a user message to exactly the sub-agents it needs, with every required role present and none spurious. \textbf{Analysis} answers a natural-language question over the replayed trace. It succeeds when the reported number $\hat{v}$ matches the value $v$ of an independent reference query, $|\hat{v}-v|\le\max(0.02\,|v|,\,0.5)$, and is backed by a real query or code run. \textbf{Memory} executes a write instruction. It succeeds when the resulting user profile and proactive task state match the request, with the protected file header intact and no collateral edits. \textbf{Plan} drives onboarding. It succeeds when the created scheduled tasks carry valid cron expressions, cover every stated health goal and ship with a published report. \textbf{Halluc.} captures the failure mode shared across these roles, the rate of asserting a number with no backing query, measured on the analysis tasks. Each role keeps its own oracle, so a failure can be attributed to the role that caused it, a diagnosis unavailable in a single-role agent design.

\begin{table*}[t]
\centering
\setlength{\tabcolsep}{5pt}
\begin{tabular}{@{}lccccccccr@{}}
\toprule
Backbone & Route & Analysis & Memory & Plan & Halluc.$\downarrow$ & MT & S5 & tok/sess & s/sess$\downarrow$ \\
\midrule
Gemini-3.5-flash$^\dagger$   & 0.98 & \textbf{0.94} & 0.89 & 0.98 & 0.01 & \textbf{0.98} & 0.37 & 42k & -- \\
DeepSeek-V4-flash$^\dagger$  & 1.00 & 0.85 & 0.97 & \textbf{1.00} & 0.00 & 0.84 & 0.00 & 66k & -- \\
GPT-5-mini$^\dagger$         & 0.88 & 0.76 & 0.99 & 0.88 & 0.00 & 0.63 & 0.00 & 206k & -- \\ \midrule
Qwen3.6-27b                  & 1.00 & 0.91 & \textbf{1.00} & 0.99 & 0.00 & 0.76 & 0.20 & 253k & 97.3 \\
Qwen3.5-35b-a3b              & 1.00 & 0.91 & 0.98 & 0.84 & 0.00 & 0.96 & 0.26 & 169k & 21.3 \\
Qwen3-32b                    & 1.00 & 0.68 & 0.81 & 0.80 & 0.00 & 0.69 & 0.00 & 178k & 55.9 \\
Gemma-4-26b-a4b              & 1.00 & 0.48 & 0.93 & 0.33 & 0.26 & 0.89 & \textbf{0.44} & 236k & 32.8 \\
Qwen3-8b                     & 0.61 & 0.47 & 0.96 & 0.57 & 0.00 & 0.68 & 0.00 & 194k & 45.6 \\
Qwen3-4b                    & 1.00 & 0.43 & 0.67 & 0.37 & 0.00 & 0.57 & 0.00 & 33k & 12.3 \\
Qwen3-1.7b                  & 0.50 & 0.41 & 0.01 & 0.99 & 0.00 & 0.26 & 0.00 & 123k & 10.8 \\
Qwen3-30b-a3b                & 0.97 & 0.36 & 0.33 & 0.99 & 0.00 & 0.61 & 0.00 & 242k & 34.7 \\
Qwen2.5-1.5b                 & 0.37 & 0.00 & 0.00 & 0.00 & 0.47 & 0.26 & 0.00 & 20k & \phantom{0}4.0 \\
\bottomrule
\end{tabular}
\caption{Per-role capability for 12 representative backbones, 3 hosted ($\dagger$) and 9 local, auto-scored on replayed public traces as pass@1 over three runs. MT is the mean per-turn score and S5 the strict all-five-turns rate of the composite rundown, Halluc.\ the unbacked-number rate, tok/sess the mean tokens per five-turn session, and s/sess the median wall-clock seconds of that session on one isolated H200 (Appendix~\ref{app:latency}), shown as -- for the hosted APIs, which we do not self-serve. The full 22-backbone sweep is in Table~\ref{tab:eval-roles-full}.}
\label{tab:eval-roles}
\end{table*}

\paragraph{Per-role results.} Table~\ref{tab:eval-roles} reports results for 12 representative backbones. Three findings stand out. First, local deployment rivals hosted frontier models. The strongest local backbones reach an analysis score of 0.91 (Qwen3.6-27b and Qwen3.5-35b-a3b), matching or surpassing DeepSeek-V4-flash (0.85) and GPT-5-mini (0.76) with no hallucinations; only Gemini-3.5-flash clearly leads (0.94).
Second, capability does not scale monotonically with model size. Qwen3-30b-a3b, despite $\sim$3B active of 30B total parameters, trails the smaller dense Qwen3-8b on analysis (0.36 versus 0.47) and memory (0.33 versus 0.96). Analysis is the most discriminative capability, ranging 0.00--0.94, while routing saturates for the most capable backbones.
Third, small models lag substantially. The 1.5B model fails across all capabilities with the highest hallucination rate (0.47), so current small on-device models remain insufficient for personal health agent tasks.

\paragraph{Composite rundown.} The per-role scores measure each capability in isolation, which overstates what a user experiences. We therefore run a \emph{composite rundown}, a five-turn persistent chat ordered as a realistic daily interaction, including a weekly check-in plan covering sleep and activity, recalling what was just set up, the average heart rate on the user's most active day (requiring a grounded number), a durable preference written to memory, and a final turn that compounds the recall, switches to metric units, and asks for the all-time peak heart rate, so both the number and the preference must be correct.
With $a_{i,t}$ and $r_{i,t}$ marking whether turn $t$ of session $i$ passes its task and routing checks, over $N$ sessions we report MT (Mean per Turn) score and S5 (Strict all-five-turns pass) score,
\begin{equation*}
\mathrm{MT} = \frac{1}{5N}\sum_{i,t}a_{i,t}, \quad
\mathrm{S5} = \frac{1}{N}\sum_{i}\prod_{t}r_{i,t}\,a_{i,t}.
\end{equation*}
We also report s/sess, the median wall-clock of one session on a single isolated H200, the latency a user waits per rundown (Appendix~\ref{app:latency}).
Performance degrades over long interaction trajectories. Per-turn scores remain strong, with Gemini-3.5-flash at 0.98 and Qwen3.5-35b-a3b at 0.96, yet flawless sessions are rare because per-turn errors compound. The best S5 values are only 0.44 (Gemma-4-26b-a4b) and 0.37 (Gemini-3.5-flash), and most backbones sit at zero. Reliably chaining competent roles across a session is therefore an open frontier for on-device agents.

\paragraph{Does decomposition help?} Splitting the agent into roles is first an engineering choice, as each role owns a bounded slice of the task, easing management and attributing failures rather than tangling tool calls inside a single persona. The ablation shows this split is also nearly free on the metric (Table~\ref{tab:eval-ablation}). Most backbones that can run the multi-role protocol gain or break even, led by $+0.65$ for Qwen3-8b and $+0.23$ for Qwen3-32b, with only Qwen3.6-27b ($-0.36$) and Gemma-4-e4b ($-0.71$) paying a substantial cost. In deployment, where a monolithic persona must hold every role in one long context, the advantage of factored responsibilities should only widen.

\begin{figure}[t]
\centering
\includegraphics[width=0.99\columnwidth]{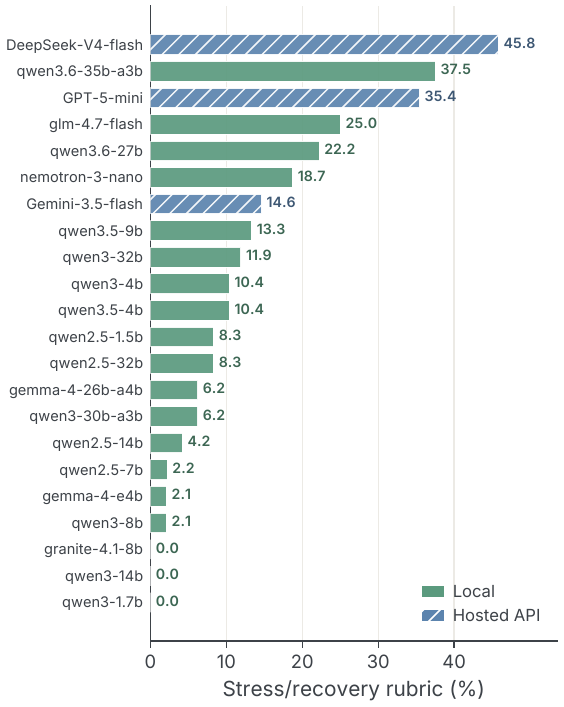}
\caption{Cross-modal report understanding, the stress/recovery rubric score (0 to 100\%) of a wearable-only summary judged against the participant's held-out self-report. The breakdown is in Table~\ref{tab:eval-report-full}.}
\label{fig:report}
\end{figure}

\subsection{Data-to-report evaluation}
\label{sec:eval:report}

When asked for a personalised health report, the agent must read a participant's multivariate history and commit to what it means. This cross-modal task, from time series to a natural-language verdict, probes understanding rather than retrieval~\citep{liu-etal-2026-ddr}. A fixed LLM judge~\citep{liu-etal-2023-geval} scores the wearable-only summary against the participant's held-out self-report on a rubric covering fatigue, readiness and stress. Figure~\ref{fig:report} shows the task is hard. The best backbone reaches only 46\% (DeepSeek-V4-flash), and Gemini-3.5-flash leads single-turn analysis at 0.94 yet scores 14.6\% here, while nemotron-3-nano at 0.85 analysis reaches 18.7\%. Narrating subjective state is therefore a competence separate from retrieving numbers, the one a personal agent adds on top of question answering.

\subsection{Real-time triggering analysis}
\label{sec:eval:trigger}

An always-on health agent must act unprompted, monitoring a continuous wearable stream without paying for an LLM call at every reading. \hime{} lets the agent compile a user's intent into a trigger, a cheap statistical condition that runs continuously for free and wakes the expensive multi-turn analysis only when it fires (\Sref{sec:arch}). To test whether this gating preserves detection while cutting cost, we compare it against a polling baseline that runs the identical analysis on a fixed schedule, measuring how well each discovers injected anomalies in the stream (Appendix~\ref{app:trigger} gives the full protocol).
Table~\ref{tab:eval-trigger-detect} reports conditional recall given trigger activation for four backbones. Agent-authored triggers identify more events on average than polling, and the gain is larger for smaller models, which a fixed schedule otherwise overwhelms with normal signals so that critical readings are missed. For larger models the trigger instead buys an order-of-magnitude cut in token usage at only a modest cost in recall.

\begin{table}[t]
\centering
\begin{tabular}{@{}lccc@{}}
\toprule
Backbone & Polling & Trigger & Tokens \\
 & recall & recall & saved \\
\midrule
Qwen3-1.7b      & 0.00 & 0.29 & 121$\times$ \\
Qwen3-8b        & 0.69 & 0.89 & 76$\times$ \\
Gemma-4-26b-a4b & 0.57 & 0.43 & 83$\times$ \\
Qwen3-30b-a3b   & 0.58 & 0.52 & 49$\times$ \\
\bottomrule
\end{tabular}
\caption{Detection on injected anomalies across PMData, LifeSnaps and MMASH, five participants per corpus and 47 events per backbone.}
\label{tab:eval-trigger-detect}
\end{table}

\subsection{Longitudinal Real-User Study}
\label{sec:eval:ux}

\begin{figure}[htbp]
\centering
\includegraphics[width=0.99\columnwidth]{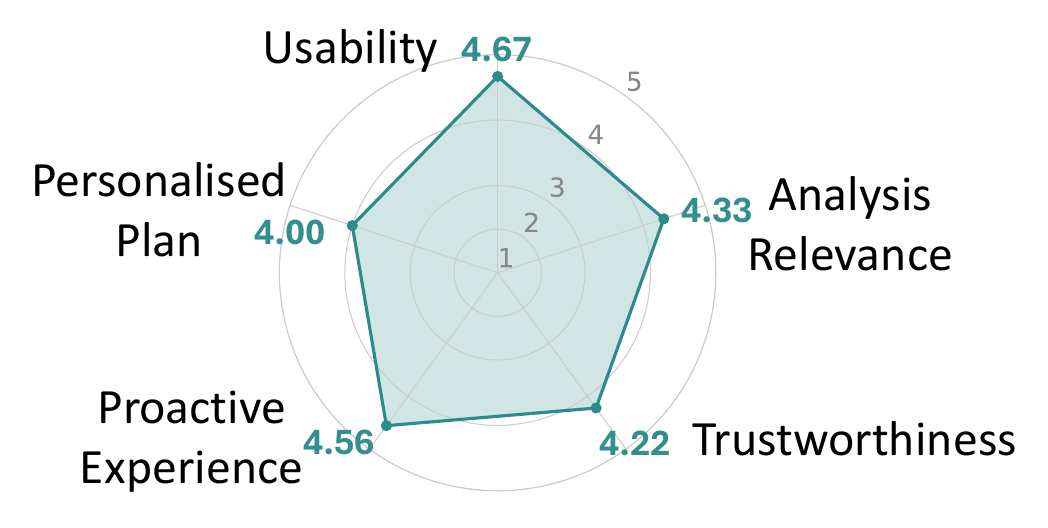}
\caption{Mean participant ratings of \hime{} across five dimensions on a five-point Likert scale.}
\label{fig:human_report}
\end{figure}

Whether \hime{} helps users is a question only real use can answer, and the aim is not momentary satisfaction but a healthier self over time. We conducted a two-month field study with nine non-technical participants using \hime{} on their own devices, collecting end-of-study feedback on a five-point Likert scale (Figure~\ref{fig:human_report}; survey instrument and further breakdown in Appendix~\ref{app:survey}).
Ratings were positive but uneven across dimensions (4.00--4.67). Usability (4.67) and proactive experience (4.56) scored highest, with participants reporting that proactive sleep and recovery reports helped them re-engage with previously overlooked health data. Personalised plan fit was the weakest dimension (4.00), and a few participants found scheduled plans slow to adapt to a changed routine. Compared with prior tools, participants credited \hime{} mainly with gains in proactivity and personalisation, and less often with gains in raw accuracy or speed (Appendix~\ref{app:survey}). Reported failures were infrequent and mostly minor, so they did not outweigh the perceived benefit of continuous, evidence-backed monitoring, though the sample remains small and self-selected.

\section{Conclusion}
We presented \hime{}, a self-hosted agent platform that turns continuous wearable streams into personal health insights, with three design principles making its architecture measurable, auditable and extensible. Across five corpora and 22 backbones, strong local models already rival hosted APIs, making privacy-preserving personal health agents increasingly practical, though multi-turn reliability and data-to-report narration remain open. We hope \hime{} supports research toward trustworthy personalised health agents.

\section*{Acknowledgements}
This work was supported in part by the UK Engineering and Physical Sciences Research Council through a Turing AI Fellowship (grant no. EP/V020579/1, EP/V020579/2) and Inkfish through the EMBRACE research programme. Wei is supported by a PhD studentship provided by King's College London (KCL). The authors acknowledge the use of Computational Research, Engineering and Technology Environment (CREATE) at KCL.

\bibliography{custom}

\clearpage
\appendix

\setcounter{table}{0}
\renewcommand{\thetable}{A\arabic{table}}
\setcounter{figure}{0}
\renewcommand{\thefigure}{A\arabic{figure}}

\section{Report Rubric Breakdown}
\label{app:report}

This appendix reports the complete result tables behind \Sref{sec:eval}, using the same replayed-trace protocol and terminal-state oracles as the main text. Table~\ref{tab:eval-report-full} breaks the data-to-report rubric of \Sref{sec:eval:report} into its three dimensions on 16 PMData participants.

\begin{table}[ht]
\centering\small
\renewcommand{\arraystretch}{0.8}
\resizebox{\columnwidth}{!}{%
\begin{tabular}{@{}lcccc@{}}
\toprule
Backbone & Overall\% & Fatigue\% & Readi.\% & Stress\% \\
\midrule
DeepSeek-V4-flash$^\dagger$ & 45.8 & 62.5 & 31.2 & 43.8 \\
Qwen3.5-35b-a3b             & 37.5 & 25.0 & 18.8 & 68.8 \\
GPT-5-mini$^\dagger$        & 35.4 & 37.5 & 18.8 & 50.0 \\
glm-4.7-flash               & 25.0 & 37.5 & 25.0 & 12.5 \\
Qwen3.6-27b                 & 22.2 & 20.0 & 20.0 & 26.7 \\
nemotron-3-nano             & 18.7 & 37.5 & \phantom{0}6.2 & 12.5 \\
Gemini-3.5-flash$^\dagger$  & 14.6 & \phantom{0}6.2 & 18.8 & 18.8 \\
Qwen3.5-9b                  & 13.3 & 20.0 & \phantom{0}0.0 & 20.0 \\
Qwen3-32b                   & 11.9 & \phantom{0}7.1 & 14.3 & 14.3 \\
Qwen3-4b                    & 10.4 & \phantom{0}6.2 & 18.8 & \phantom{0}6.2 \\
Qwen3.5-4b                  & 10.4 & 12.5 & \phantom{0}6.2 & 12.5 \\
Qwen2.5-1.5b                & \phantom{0}8.3 & \phantom{0}6.2 & 12.5 & \phantom{0}6.2 \\
Qwen2.5-32b                 & \phantom{0}8.3 & 18.8 & \phantom{0}0.0 & \phantom{0}6.2 \\
Gemma-4-26b-a4b             & \phantom{0}6.2 & \phantom{0}6.2 & \phantom{0}6.2 & \phantom{0}6.2 \\
Qwen3-30b-a3b               & \phantom{0}6.2 & 18.8 & \phantom{0}0.0 & \phantom{0}0.0 \\
Qwen2.5-14b                 & \phantom{0}4.2 & \phantom{0}6.2 & \phantom{0}6.2 & \phantom{0}0.0 \\
Qwen2.5-7b                  & \phantom{0}2.2 & \phantom{0}0.0 & \phantom{0}0.0 & \phantom{0}6.7 \\
Gemma-4-e4b                 & \phantom{0}2.1 & \phantom{0}0.0 & \phantom{0}0.0 & \phantom{0}6.2 \\
Qwen3-8b                    & \phantom{0}2.1 & \phantom{0}0.0 & \phantom{0}0.0 & \phantom{0}6.2 \\
granite-4.1-8b              & \phantom{0}0.0 & \phantom{0}0.0 & \phantom{0}0.0 & \phantom{0}0.0 \\
Qwen3-14b                   & \phantom{0}0.0 & \phantom{0}0.0 & \phantom{0}0.0 & \phantom{0}0.0 \\
Qwen3-1.7b                  & \phantom{0}0.0 & \phantom{0}0.0 & \phantom{0}0.0 & \phantom{0}0.0 \\
\bottomrule
\end{tabular}}
\caption{Full cross-modal report results on 16 PMData participants: overall rubric percentage and per-dimension hit rate. $\dagger$ hosted API.}
\label{tab:eval-report-full}
\end{table}

\begin{table}[t]
\centering\small
\renewcommand{\arraystretch}{0.8}
\resizebox{1.0\columnwidth}{!}{%
\begin{tabular}{@{}lcccccccc@{}}
\toprule
Backbone & Rt & An & Mem & Pl & H$\downarrow$ & MT & S5 & tok \\
\midrule
Gemini-3.5-flash$^\dagger$      & 0.98 & 0.94 & 0.89 & 0.98 & 0.01 & 0.98 & 0.37 & 42k \\
Qwen3.6-27b                 & 1.00 & 0.91 & 1.00 & 0.99 & 0.00 & 0.76 & 0.20 & 253k \\
Qwen3.5-35b-a3b             & 1.00 & 0.91 & 0.98 & 0.84 & 0.00 & 0.96 & 0.26 & 169k \\
Qwen3.5-9b                  & 1.00 & 0.87 & 1.00 & 1.00 & 0.00 & 0.91 & 0.01 & 44k \\
nemotron-3-nano             & 0.99 & 0.85 & 0.93 & 0.66 & 0.00 & 0.02 & 0.00 & 52k \\
DeepSeek-V4-flash$^\dagger$ & 1.00 & 0.85 & 0.97 & 1.00 & 0.00 & 0.84 & 0.00 & 66k \\
Qwen3.5-4b                  & 1.00 & 0.77 & 0.84 & 0.18 & 0.00 & 0.66 & 0.00 & 50k \\
GPT-5-mini$^\dagger$        & 0.88 & 0.76 & 0.99 & 0.88 & 0.00 & 0.63 & 0.00 & 206k \\
Qwen2.5-32b                 & 1.00 & 0.72 & 0.66 & 0.07 & 0.00 & 0.74 & 0.22 & 159k \\
Qwen3-32b                   & 1.00 & 0.68 & 0.81 & 0.80 & 0.00 & 0.69 & 0.00 & 178k \\
granite-4.1-8b              & 0.60 & 0.64 & 0.93 & 0.80 & 0.00 & 0.00 & 0.00 & 34k \\
Qwen3-14b                   & 1.00 & 0.52 & 0.79 & 0.00 & 0.00 & 0.44 & 0.00 & 162k \\
Gemma-4-26b-a4b             & 1.00 & 0.48 & 0.93 & 0.33 & 0.26 & 0.89 & 0.44 & 236k \\
Qwen3-8b                    & 0.61 & 0.47 & 0.96 & 0.57 & 0.00 & 0.68 & 0.00 & 194k \\
Qwen3-4b                    & 1.00 & 0.43 & 0.67 & 0.37 & 0.00 & 0.57 & 0.00 & 33k \\
Qwen3-1.7b                  & 0.50 & 0.41 & 0.01 & 0.99 & 0.00 & 0.26 & 0.00 & 123k \\
Qwen2.5-7b                  & 0.60 & 0.38 & 0.66 & 0.91 & 0.12 & 0.31 & 0.00 & 38k \\
Qwen3-30b-a3b               & 0.97 & 0.36 & 0.33 & 0.99 & 0.00 & 0.61 & 0.00 & 242k \\
glm-4.7-flash               & 1.00 & 0.27 & 0.83 & 0.92 & 0.00 & 0.00 & 0.00 & 33k \\
Qwen2.5-14b                 & 0.74 & 0.15 & 0.63 & 0.22 & 0.22 & 0.62 & 0.00 & 102k \\
Gemma-4-e4b                 & 1.00 & 0.02 & 0.67 & 0.00 & 0.00 & 0.37 & 0.00 & 144k \\
Qwen2.5-1.5b                & 0.37 & 0.00 & 0.00 & 0.00 & 0.47 & 0.26 & 0.00 & 20k \\
\bottomrule
\end{tabular}}
\caption{Full per-role capability over 22 backbones (pass@1, three runs), sorted by analysis; abbreviated columns as in Table~\ref{tab:eval-roles}. $\dagger$ hosted API.}
\label{tab:eval-roles-full}
\end{table}

\begin{table}[!t]
\centering
\setlength{\tabcolsep}{6pt}
\begin{tabular}{@{}lccc@{}}
\toprule
Backbone & Dec & Mono & $\Delta$ \\
\midrule
Qwen3-8b        & 0.98 & 0.33 & $+0.65$ \\
Qwen3-32b       & 0.52 & 0.29 & $+0.23$ \\
Qwen2.5-7b      & 0.21 & 0.09 & $+0.12$ \\
Qwen3-14b       & 0.77 & 0.66 & $+0.11$ \\
Qwen3-30b-a3b   & 0.29 & 0.22 & $+0.07$ \\
Gemma-4-26b-a4b & 0.69 & 0.63 & $+0.06$ \\
Qwen3-1.7b      & 0.00 & 0.00 & $\phantom{+}0.00$ \\
Qwen2.5-1.5b    & 0.00 & 0.00 & $\phantom{+}0.00$ \\
Qwen3.5-35b-a3b & 0.90 & 0.91 & $-0.01$ \\
Qwen2.5-32b     & 0.07 & 0.11 & $-0.04$ \\
Qwen2.5-14b     & 0.80 & 0.89 & $-0.09$ \\
Qwen3.6-27b     & 0.63 & 0.99 & $-0.36$ \\
Gemma-4-e4b     & 0.01 & 0.72 & $-0.71$ \\
\bottomrule
\end{tabular}
\caption{Decomposition ablation, decomposed (Dec) versus monolithic (Mono), for the 13 local backbones in the ablation run, sorted by $\Delta$.}
\label{tab:eval-ablation}
\end{table}

\section{Full Per-Role Sweep}
\label{app:roles}

Table~\ref{tab:eval-roles-full} extends Table~\ref{tab:eval-roles} of \Sref{sec:eval:roles} to all 22 backbones, sorted by analysis score, and sharpens the main findings. The newest local generations close most of the gap to hosted APIs, with Qwen3.5-9b the cheapest competent configuration, reaching 0.87 analysis with perfect memory and plan scores at only 44k tokens. Single-turn skill does not guarantee session-level reliability. Nemotron-3-nano pairs 0.85 analysis with a collapsed multi-turn score of 0.02, and routing saturates near 1.00 for over half the field yet falls to 0.37 at the 1.5B floor.

\section{Decomposition Ablation}
\label{app:ablation}

Table~\ref{tab:eval-ablation} reports the single-turn composed task of \Sref{sec:eval:roles} for the 13 local backbones of the ablation run (the other nine postdate this run). The full list bears out the main text. Six backbones gain from the split, led by Qwen3-8b at $+0.65$ and Qwen3-32b at $+0.23$, and five more sit within $0.09$ of break-even. Only Qwen3.6-27b at $-0.36$ and Gemma-4-e4b at $-0.71$ pay a real cost, while the two smallest fail under either protocol. Factoring into roles therefore costs little even on a single turn, the setting least favourable to it.

\section{Trigger Detection Protocol}
\label{app:trigger}

This appendix details the detection experiment of \Sref{sec:eval:trigger}. The \emph{same} agent invoked two ways, \textbf{polling} (full multi-turn analysis on a fixed five-minute schedule) versus \textbf{trigger} (identical analysis only when an agent-authored condition fires). The backbone authors its own rules from one plain-language request covering spikes, sustained elevation, slow heart rate, low blood oxygen and low heart-rate variability, each choosing a signal, a threshold or standard-deviation condition, a window and a cooldown. We replay PMData, LifeSnaps and MMASH, the three intraday heart-rate corpora, under a frozen clock that ingests batches in order so a rule only sees samples up to the simulated present.
Since no corpus labels real anomalies, we inject a fixed, diverse set of events as ground truth. Heart-rate spikes and drops (baseline mean $\pm$ several standard deviations, floored or ceilinged at a fixed bpm), a sustained tachycardic run, a blood-oxygen dip to 86\%, and a low heart-rate-variability sample, each replacing the real reading in its batch so it is the latest value the signal has when the gate evaluates. A fire is credited only within a 45-minute grace window after onset, and a fixed LLM judge~\citep{liu-etal-2023-geval} checks, per event, whether the woken analysis surfaced that abnormality, flagging unsupported concerns as false alarms. We report recall on events whose gate fires, over five participants per corpus and 47 events per backbone; the token saving in Table~\ref{tab:eval-trigger-detect} is the ratio of the trigger arm's few daily analyses to polling's fixed schedule.

\section{On-Device Latency Profile}
\label{app:latency}

The main text reports token count because it is hardware-independent, but \hime{} runs on the user's own machine, so wall-clock latency matters too. Table~\ref{tab:eval-latency} reports the \emph{rundown} for every local backbone, the end-to-end wall-clock of the five-turn composite session of \Sref{sec:eval:roles} (the session scored there for MT and S5), each served alone on one NVIDIA H200 through vLLM~\citep{kwon-etal-2023-vllm} in bfloat16 (bf16) with reasoning backbones thinking, median over five PMData participants. Decode throughput at batch one ranged from 53 tok/s at 32B to 536 tok/s at 1.5B, tracking active rather than total parameters, so the 30B mixture-of-experts decoded as fast as a dense 4B.

\begin{table}[t]
\centering\small
\setlength{\tabcolsep}{8pt}
\renewcommand{\arraystretch}{0.99}
\begin{tabular}{@{}lrr@{}}
\toprule
Backbone & Params & Rundown$\downarrow$ (s) \\
\midrule
Qwen2.5-1.5b      & 1.5B          & 4.0 \\
glm-4.7-flash     & 31B$^\dagger$ & 4.3 \\
Qwen2.5-7b        & 7B            & 9.4 \\
Qwen3-1.7b        & 1.7B          & 10.8 \\
nemotron-3-nano   & 4B            & 12.2 \\
Qwen3-4b          & 4B            & 12.3 \\
Gemma-4-e4b       & 8B            & 12.4 \\
Qwen3.5-35b-a3b   & 35B$^\dagger$ & 21.3 \\
Qwen3.5-9b        & 9B            & 23.8 \\
Qwen2.5-14b       & 14B           & 26.7 \\
Qwen3.5-4b        & 4B            & 27.9 \\
Qwen3-14b         & 14B           & 29.4 \\
Gemma-4-26b-a4b   & 26B$^\dagger$ & 32.8 \\
Qwen3-30b-a3b     & 30B$^\dagger$ & 34.7 \\
Qwen2.5-32b       & 32B           & 44.6 \\
Qwen3-8b          & 8B            & 45.6 \\
Qwen3-32b         & 32B           & 55.9 \\
Qwen3.6-27b       & 27B           & 97.3 \\
granite-4.1-8b    & 8B            & 230.2$^\ddagger$ \\
\bottomrule
\end{tabular}
\caption{Composite-rundown latency on one NVIDIA H200, median over five PMData participants. $^\dagger$ mixture-of-experts. $^\ddagger$ unreliable: loops on malformed tool calls until timeout.}
\label{tab:eval-latency}
\end{table}

\section{Human Evaluation Survey Design}
\label{app:survey}

\begin{figure}[ht]
\centering
\includegraphics[width=0.99\columnwidth]{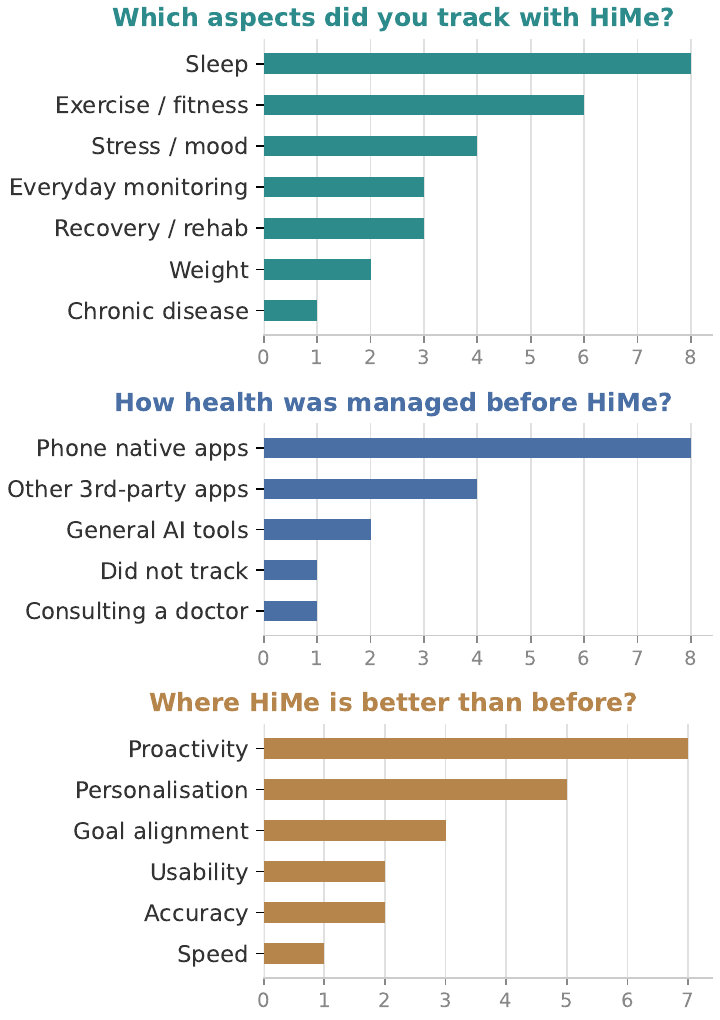}
\caption{Participants' multiple-choice responses on \hime{} usage scenarios, prior health-management methods, and advantages over previous approaches.}
\label{fig:human_report_bar}
\end{figure}

This appendix gives the survey instrument behind \Sref{sec:eval:ux}. Beyond the Likert questionnaire (Figure~\ref{fig:human_report}), participants answered three multiple-choice questions, including which health domains they tracked, how they managed those before \hime{}, and where \hime{} improved on their previous approach (Figure~\ref{fig:human_report_bar}).
Participants adopted \hime{} as a daily sleep companion, extending to fitness and stress management, having relied mostly on fragmented apps before. They credited \hime{} with proactivity and personalisation gains, fewer in usability, accuracy or speed (\Sref{sec:eval:ux}).

\end{document}